\documentclass[letterpaper, 10 pt, conference]{ieeeconf}  

\IEEEoverridecommandlockouts                              

\usepackage{graphics} 
\usepackage{epsfig} 
\usepackage{mathptmx} 
\usepackage{times} 
\usepackage{amsmath} 
\usepackage{amssymb}  
\usepackage{optidef}
\usepackage{algorithm,algorithmic}

\usepackage{cite}
\usepackage{multirow}
\usepackage{color}
\usepackage{amsmath,amssymb,amsfonts}
\usepackage{balance}
\usepackage{bm}
\usepackage{hyperref}

\makeatletter
\let\NAT@parse\undefined
\makeatother

\usepackage{hyperref}

\usepackage{caption}
 \definecolor{somegray}{rgb}{0.5, 0.5, 0.5}
 \newcommand{\darkgrayed}[1]{\textcolor{somegray}{#1}}
 \makeatletter
 \newcommand*\titleheader[1]{\gdef\@titleheader{#1}}
 \AtBeginDocument{%
   \let\st@red@title\@title
   \def\@title{%
     \vskip-3em
     \bgroup\normalfont\large\centering\@titleheader\par\egroup
     \vskip1.1em\st@red@title}
 }

\titleheader{\darkgrayed{This paper has been accepted for publication at the \\ IEEE International Conference on Robotics and Automation (ICRA), London, 2023. \copyright IEEE}}

\title{\LARGE \bf
 Obstacle Identification and Ellipsoidal Decomposition for Fast Motion Planning in Unknown Dynamic Environments 
}

\author{Mehmetcan Kaymaz and Nazım Kemal Ure
\thanks{Mehmetcan Kaymaz is with the Faculty of Aeronautics and Astronautics,
        Istanbul Technical University, Istanbul, Turkey
        {\tt\small kaymazm16@itu.edu.tr}}%
\thanks{Nazim Kemal Ure is with the Faculty of Computer and Informatics Engineering, Istanbul Technical University, Istanbul, Turkey
        {\tt\small ure@itu.edu.tr}}%
}

\begin{document}

\maketitle
\thispagestyle{empty}
\pagestyle{empty}

\begin{abstract}

Collision avoidance in the presence of dynamic obstacles in unknown environments is one of the most critical challenges for unmanned systems. In this paper, we present a method that identifies obstacles in terms of ellipsoids to estimate linear and angular obstacle velocities. Our proposed method is based on the idea of any object can be approximately expressed by ellipsoids. To achieve this, we propose a method based on variational Bayesian estimation of Gaussian mixture model, the Kyachiyan algorithm, and a refinement algorithm. Our proposed method does not require knowledge of the number of clusters and can operate in real-time, unlike existing optimization-based methods. In addition, we define an ellipsoid-based feature vector to match obstacles given two timely close point frames. Our method can be applied to any environment with static and dynamic obstacles, including ones with rotating obstacles. We compare our algorithm with  other clustering methods and show that when coupled with a trajectory planner, the overall system can efficiently traverse unknown environments in the presence of dynamic obstacles.
\end{abstract}

\section{INTRODUCTION}
Obstacle avoidance in unknown environments is a common problem for unmanned systems. In general, the identification of obstacles and trajectory planning are considered separate problems, and the computational complexity of both algorithms is desired to be as low as possible\cite{CHEN2022104124} to compensate for the uncertainty of the environment. In scenarios that involve dynamic obstacles, the identification of obstacles has a critical effect on motion estimation, as the estimation performance strongly depends on the accurate identification of such obstacles.

Clustering is a popular method for separating obstacles based on point cloud data provided by perception systems. The number of estimated clusters can make a big difference in the overall performance. There are several approaches to compute the number of clusters; such as assuming the constant number of clusters\cite{Esrafilian2016}\cite{Mok2017}, calculating the number of clusters from the number of points\cite{Corah2019}, or estimating the number of clusters based on point data\cite{David2006}\cite{Ester1996}. Most of the previous work assumes that each dynamic obstacle can be expressed as a single circle\cite{Lin2020} or an ellipsoid\cite{Brito2019}\cite{falanga2020} and moves without rotation\cite{CHEN2022104124}. It has been demonstrated that the ellipsoid shape is a better approximation to approximate obstacles in such environments\cite{Chakravarthy2011}. Although these assumptions are true for a lot of real-world applications, these methods face severe limitations as the environment structure gets more complex.

In this work, we model the obstacle identification problem as the computation of minimum volume ellipsoids\cite{Shioda2007}, which is defined as finding $n$ ellipsoids that cover $m$ given points where the total volume of ellipsoids is minimized. The optimal solution can be obtained via mixed-integer semidefinite programming(MISDP). However, MISDP formulation requires providing the number of ellipsoids beforehand, and the computational complexity of solving MISDP is not feasible for online planning. We propose a method that provides an approximate solution for the minimum volume ellipsoids problem without knowing the number of ellipsoids a priori. In addition, the computational requirement of our proposed algorithm is much lesser than MISDP, which enables running trajectory planning algorithms concurrently with obstacle identification. 

In summary, the main contributions of the paper are as follows:
\begin{itemize}
    \item  To solve the minimum volume ellipsoids problem without the knowledge of the number of ellipsoids, we combine variational Bayesian estimation of Gaussian mixture model(VIGMM)\cite{David2006} with the Khachiyan algorithm\cite{Khachiyan1996} and add custom refinements to improve the performance.
    
    \item  To match obstacles given two timely close point frames, we define a feature vector based on ellipsoid parameters and estimate motion by using matched ellipsoids. 
    
    \item Overall, our proposed work can rapidly identify dynamic obstacles in cluttered environments in real time. The simulation results show that the proposed identification method can be coupled with a trajectory planning method to traverse a variety of dynamic environments without too much computational overhead.
\end{itemize}

\section{RELATED WORK}
Clustering is a common method to identify obstacles in collision avoidance approaches, especially when the obstacles are dynamic. K-means\cite{MacQueen1967SomeMF}\cite{Altman1992} is a famous method in the literature for clustering due to its low computation requirement. In \cite{Esrafilian2016}, the authors used the K-means algorithm with a fixed number of clusters to classify point cloud data from monocular SLAM in a static and simple environment. Hierarchical clustering was used with the single-linkage clustering method in \cite{PARK2020105882}, where point cloud data was estimated from the stereo vision system. A similar idea is applied in \cite{Pasqual2022} with lidar to generate point cloud data. DBSCAN\cite{Ester1996} based on Euclidean distance is a common method for obstacle avoidance especially when the point data is noisy. In \cite{falanga2020}, point cloud data obtained from the event camera was clustered using the DBSCAN algorithm for dynamic obstacle avoidance. The same clustering method is also used in \cite{CHEN2022104124} for both dynamic and static obstacles where the point cloud is estimated from the depth image. The problem with the methods mentioned above is that they fail when the clusters are Gaussian data points in which the point distribution is ellipsoidal, so Euclidian distance is not a sufficient metric to separate clusters.

The MISDP is proposed to cluster Gaussian data points in \cite{Shioda2007} but the algorithm's computational complexity is high and it requires the number of clusters as input. A method based on C-means was proposed in \cite{Xenaki2018}, which successfully clusters Gaussian data points but the computational cost is high, similar to the MISDP. The possibilistic C-means algorithm\cite{Krishnapuram1993}\cite{Krishnapuram1996}, which is the primitive version of \cite{Xenaki2018} is more computationally efficient but it fails when the clusters overlap. Also, C-means methods require the number of clusters. Another method to cluster Gaussian data points is Gaussian mixture models(GMM) which are based on fitting multivariate Gaussian distributions to data. In \cite{Mok2017}, GMM was concatenated with the artificial potential field method for static obstacle avoidance where the number of clusters is selected constant. In \cite{Dhawale2018}, online mapping based on GMM with a constant number of Gauss components is proposed for static collision avoidance. \cite{Corah2019} used GMM to generate the online map but the number of Gauss components is estimated via $N/R$ where $N$ is the number of points and $R$ is a constant parameter. 


Matching the obstacles/clusters from sequential point frame measurements is important for estimating the motion of dynamic objects. In \cite{Eppenberger2020} and \cite{Wang2021}, the centers of the clusters were used to match obstacles whereas the ones with a minimum distance between center points are assumed as the same obstacle. However, this method may fail when the centers of two different obstacles are close to each other. To improve the matching robustness, the feature vector is designed in \cite{CHEN2022104124} which is more robust than other alternatives but requires RGB vision inputs.  

\section{PROBLEM DEFINITION}
In this section, we present a minimum volume ellipsoids problem to cover a given point of data. The problem is defined in two dimensional (2D) plane.

We use three different ellipsoid definitions. The first one is called the general form, which norm based definition: 
\begin{equation}\label{generalform}
    \mathcal{E}=\{x \textbf{ } | \textbf{ } ||Ax+b||_2\leq 1 \textbf{ , } x \in \mathbb{R}^2 \},
\end{equation}
where $A \in \mathbb{S}^2_{++} $ and $b \in \mathbb{R}^2$. The second definition is called the quadratic form, which is the open version of the general form:
\begin{equation}\label{quadraticform}
    \mathcal{E}=\{x \textbf{ } | \textbf{ } x^T A^q x + 2(b^q)^T x + c^q \leq 0 \textbf{ , } x \in \mathbb{R}^2\},
\end{equation}

where $A^q \in \mathbb{S}^2_{++}$, $b^q \in \mathbb{R}^2$ and $c^q \in \mathbb{R}$. The last definition is called the standard form, which is 
a modified version of the quadratic form:
\begin{equation}\label{standartform}
    \mathcal{E}=\{x \textbf{ } | \textbf{ } (x-x_{c})^T R(\theta)^T H(r_1,r_2) R(\theta) (x-x_{c}) \leq 1 \textbf{ , } x \in \mathbb{R}^2 \},
\end{equation}
where $x_c \in \mathbb{R}^2$ is the center of the ellipsoid, $R(\theta) \in \mathbb{R}^{2 \times 2}$ is the rotation matrix, $\theta \in [0,\pi]$ is the rotation angle, $H(r_1,r_2) \in \mathbb{S}^2_{++}$ is the ellipsoidal zone, $r_1 \in \mathbb{R}^+$ and $r_2 \in \mathbb{R}^+$ are the minor
and major axes of the ellipsoid.


Let the total number of ellipsoids be given as $n_{ell} \in \mathbb{N}^+$, hence the space covered by all ellipsoids is defined as $\mathcal{E}_{all}=\mathcal{E}_1 \cup ... \cup \mathcal{E}_{n_{ell}} \subset \mathbb{R}^2$. Let the shape of the obstacle be represented by $n_{obs} \in \mathbb{N}^+$ number of points and let $K=\{1,2,...,n_{obs}\}$. The position of $k$th point is denoted as $\nu_{k} \in \mathbb{R}^2$. The set of all points in given shape is donated as $O_{obs}=\{\nu_{1},\nu_{2},...,\nu_{K}\} \subset \mathbb{R}^{2}$. Given points are covered if and only if all points in the shape are the elements of at least one ellipsoid. In other words, the set $O_{obs}$ must be a subset of $\mathcal{E}_{all}$.


\section{METHOD}

Our proposed framework is composed of two sub-modules: first, we present a method for solving the minimum volume ellipsoids problem by using variational Bayesian estimation of the Gaussian mixture model, the Khachiyan algorithm, and our refinement algorithm. Then, we show how to match obstacles from given two frames measured in a short time interval and how to estimate the motion of the obstacles via ellipsoids.

\subsection{Minimum Volume Ellipsoids}

Our approximate method starts with the Gaussian mixture model(GMM), which is used as the clustering method. The GMM is a probabilistic model with multiple Gaussian components. Given a set of points $\nu$, the probability density function of a GMM can be expressed as:

\begin{equation}
    p(\nu|\pi,\mu,\Sigma)=\sum_{j=1}^J \pi_j \mathcal{N}(\nu;\mu_j,\Sigma_j),
\end{equation}

where $J$ is the number of clusters, $\mathcal{N}(\nu;\mu_j,\Sigma_j)$ is multivariate Gaussian distribution, $\mu_j$ is mean vector, $\Sigma_j$ is covariance matrix and $\pi_j$ is the weight coefficient of the $j$th component, satisfying $\sum \pi_j=1$. 

The latent variables $\lambda=\{\lambda_1,...,\lambda_J\}$ can be estimated efficiently via the Expectation-Maximization(EM)\cite{Hjelm2018} method, but the number of Gaussian components must be determined manually. The Maximum Likelihood(ML)\cite{Bekker2020} method can be applied, but computing the global optimum is difficult. The Monte-Carlo Markov Chain(MCMC)\cite{LI2020105950} method addresses both of the issues in EM and ML, but the computational requirement is too big. On the other hand, Variational Inference(VI)\cite{WU201832} estimates the parameters of GMM through optimization, which is usually faster than MCMC. In VI, the maximum number of Gaussian components must be determined, the method eliminates the unnecessary ones.

The problem of estimating GMM parameters can be expressed as finding a posterior distribution $p(\lambda|\nu)$. In VI, a variational distribution $q(\lambda)$ is used to approximate $p(\lambda|\nu)$ by minimizing Kullback-Leible(KL) divergence between them. It is not possible to calculate $KL(q(\lambda)||p(\lambda|\nu))$ since $p(\lambda|x)$ is unknown. However, the following equation can be written using the Bayes rule:
\begin{equation}
  KL(q(\lambda)||p(\lambda|\nu))=\mathcal{L}(q)-\int_\lambda q(\lambda) ln p(\nu) d\lambda,
\end{equation}
where $\mathcal{L}(q)$ is Evidence Lower Bound(ELBO). $p(\nu)$ is a definite value, so minimizing KL diverge is equivalent to maximizing $L(q)$.

\begin{figure}
    \centering
    \includegraphics[width=.35\textwidth]{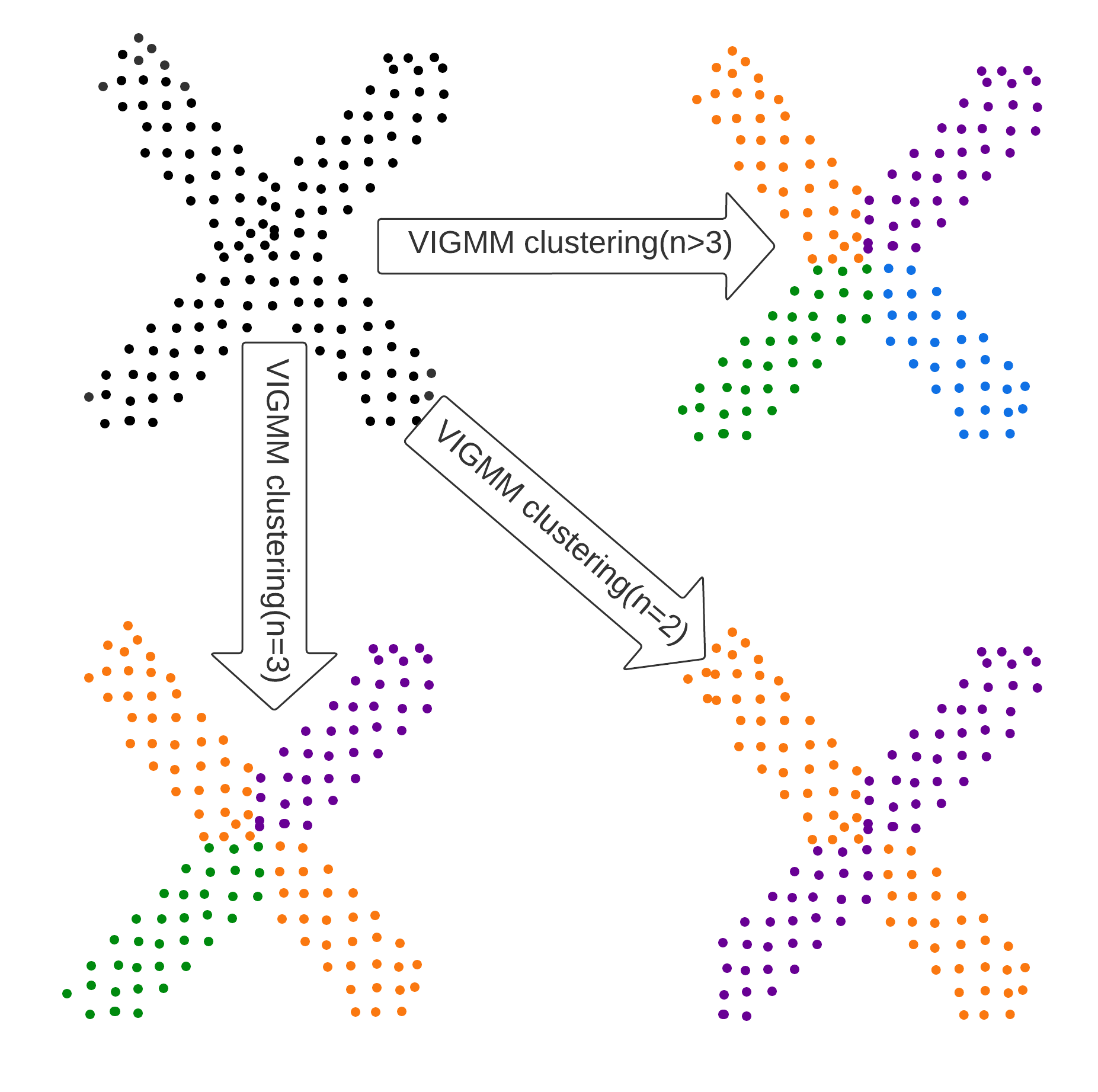}
    \caption{Variational Bayesian estimation of Gaussian mixture model clustering based on the number of maximum components. Each color shows a different cluster and n is the maximum number of clusters.}
    \label{fig:vigmm}
\end{figure}

After clustering, the minimum volume ellipsoids problem is converted to minimum volume ellipsoid for $J$ clusters which is much easier to calculate. The minimum volume ellipsoid problem is defined as finding the minimum volume ellipsoid that covers given $n_{obs}$ points which can be written in semi-definite programming form\cite{boyd2004}:

\begin{mini}[2] 
{A,b}{(\det A)^{\frac{1}{n}}}
{}{}\label{sdp}
\addConstraint{||A\nu+b||_2\leq 1}{},
\end{mini}

where $A$ and $b$ represent ellipsoid in the general form. The Khachiyan algorithm\cite{Khachiyan1996} is used for solving \eqref{sdp}, which provides a solution with a predefined error. The Kyachiyan algorithm is faster than semi-definite programming solvers even with a very small predefined error\cite{TODD20071731}. The issue in the variational Bayesian estimation of the Gaussian mixture model is observed in \autoref{fig:vigmm}. Although the VI method eliminates unnecessary Gaussian components, the result depends on the number of maximum components. The refinement algorithm uses the Khachiyan algorithm outputs, which are $n_{ell}$ ellipsoids as input and combines a subset of them based on their volumes.   

Assume we have $n_{ell}$ ellipsoids and $J=\{1,2,...,n_{ell}\}$. We define ellipsoids volume ratio $r_{ell_{i,j}}$ as:
\begin{equation}\label{eq:ratio}
    r_{ell_{i,j}}=\frac{Vol(\mathcal{E}_i)+Vol(\mathcal{E}_j)}{Vol(OBB(\mathcal{E}_i,\mathcal{E}_j))},
\end{equation}
where $i,j \in J$, $Vol$ is the volume and $OBB(\mathcal{E}_i,\mathcal{E}_j)$ is oriented bounding box of $\mathcal{E}_i$ and $\mathcal{E}_j$. Calculation of $r_{ell_{i,j}}$ is computationally efficient as the volume of $\mathcal{E}_i$ and $\mathcal{E}_j$ are already known, and calculating OBB is a simple operation\cite{Gottschalk1997}. In order to combine two ellipsoids, we use the minimum volume ellipsoid covering the union of ellipsoids\cite{boyd2004}, which is a convex optimization problem that can be solved efficiently for two ellipsoids in two or three dimensions. The semi-definite programming for given two ellipsoids in the quadratic form is defined as:
\begin{mini}[3] 
{A^2,\Tilde{b},\tau_1,\tau_2}{(\det A)^{\frac{1}{n}}}
{}{}\label{sdp2}
\addConstraint{\tau_i \geq 0}{}
\addConstraint{\begin{bmatrix}
    A^2-\tau_i A^q_i & \Tilde{b}-\tau_i b^q_i & 0 \\
    (\Tilde{b}-\tau_i b^q_i)^T & -1-\tau_i c^q_i & \Tilde{b}^T \\
    0 & \Tilde{b} & -A^2
\end{bmatrix}\preceq 0}{}
\addConstraint{i=1,2}{},
\end{mini}

where $\Tilde{b}=Ab$. The Mosek software\cite{mosek} is used for solving the semi-definite problem defined in \eqref{sdp2}. The higher values of $r_{ell_{i,j}}$ means combining $i$th and $j$th ellipsoids is a acceptable approximation. We define a threshold $r_{ell_{threshold}}$ to decide which ellipsoids should be combined. The refinement algorithm is presented in Alg. \ref{alg:refinement}

 \begin{algorithm}
 \caption{The Refinement Algorithm}
 \begin{algorithmic}[1]
 \renewcommand{\algorithmicrequire}{\textbf{Input:}}
 \renewcommand{\algorithmicensure}{\textbf{Output:}}
 \REQUIRE $\{\mathcal{E}_1,...,\mathcal{E}_{n_{ell_v}}\}$ list of ellipsoids, $r_{ell_{threshold}}$ threshold to combine ellipsoids.
 
 \ENSURE  $\{\mathcal{E}_1,...,\mathcal{E}_{n_{ell_r}}\}$ list of ellipsoids

  \FOR {$i = 1$ to $n_{ell_v}$}
  \FOR{$j = i+1$ to $n_{ell_v}$}
  
  \STATE  $r_{ell_{i,j}}=f(\mathcal{E}_i,\mathcal{E}_j)$ \eqref{eq:ratio}
  
  \IF {$r_{ell_{i,j}} \geq r_{ell_{threshold}}$}
            \STATE Delete $i$th and $j$th ellipsoids
            \STATE $A_{i,j},b_{i,j}=f(\mathcal{E}_i,\mathcal{E}_j)$\eqref{sdp2} 
            \STATE Add combined ellipsoid $A_{i,j},b_{i,j}$
  \ENDIF
  
  \ENDFOR
  \ENDFOR
 \RETURN   $\{\mathcal{E}_1,...,\mathcal{E}_{n_{ell_r}}\}$ 
 \end{algorithmic} 
  \label{alg:refinement}
 \end{algorithm}


The refinement algorithm can be used until the number of ellipsoids does not change. Yet in our simulations, we did not encounter any scenario that requires using the refinement algorithm more than once. 

 
  

 

\begin{figure*}[t]
    \centering
    \includegraphics[width=0.8\textwidth]{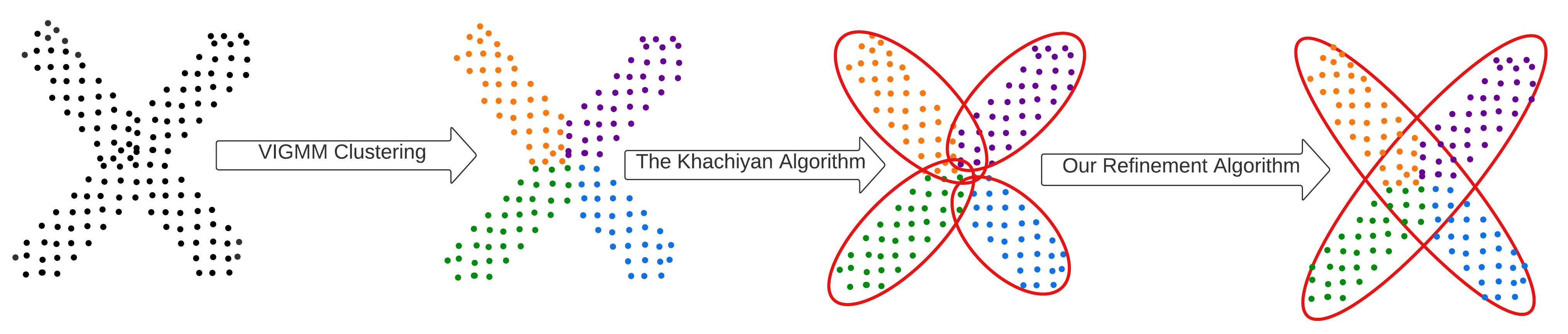} 
    \caption{Overall algorithm pipeline. Each color shows the different clusters obtained from VIGMM.}
    \label{fig:pipeline}
\end{figure*}

\subsection{Obstacle Matching and Motion Estimation}

We define the obstacle matching method and motion estimation method similar to \cite{CHEN2022104124}. The feature vector is defined as ellipsoid parameters in the standard form.
\begin{equation}
    fte(\mathcal{E})=[x_c,r_1,r_2,\theta]
\end{equation}

The idea is that if the change in future vectors given two timely close frames is small, they are considered the same object. The Euclidean distance is used to determine the change in future vectors. Assume $i$th ellipsoid generated at time $t_1$ and $j$th ellipsoid is generated at time $t_2$, then the distance between them is defined as:
\begin{equation}
    d_{i,j}=||fte(\mathcal{E}_i)-fte(\mathcal{E}_j)||_2.
\end{equation}

The $j$th ellipsoid is matched with the one from the previous frame with the smallest distance. For motion estimation, we assume that the obstacles are moving with constant linear and angular velocity. When the two ellipsoids are matched, the linear and angular velocity of $j$th ellipsoid is calculated by a simple numerical derivative.

\begin{equation}
    v_j=\frac{x_{c_j}-x_{c_i}}{t_2-t_1} \quad \omega_j=\frac{\theta_j-\theta_i}{t_2-t_1}
\end{equation}

This calculation is not directly accurate due to the fact that ellipsoids are generated with an approximation algorithm so, a Kalman filter is integrated to reduce error with more samples. 

\section{PLANNING \& CONTROL}
We use a single model predictive control algorithm for both planning and control. We use the simple point mass dynamics as the vehicle dynamics. The states $\xi=\{p_x,p_y,v_x,v_y\}$ are positions and velocities, respectively. The inputs $u=\{F_x,F_y\}$ are forces in the $x$ and $y$ directions. The state space model $\Dot{\xi}=A\xi+Bu$ is defined as follows:

\begin{equation}
    \begin{bmatrix}
        \Dot{p}_x\\
        \Dot{p}_y\\
        \Dot{v}_x\\
        \Dot{v}_y
    \end{bmatrix}=
    \begin{bmatrix}
        0&0&1&0\\
        0&0&0&1\\
        0&0&0&0\\
        0&0&0&0
    \end{bmatrix}
    \begin{bmatrix}
        p_x\\
        p_y\\
        v_x\\
        v_y
    \end{bmatrix}+
    \begin{bmatrix}
        0&0\\
        0&0\\
        1/m&0\\
        0&1/m
    \end{bmatrix}
    \begin{bmatrix}
        F_x\\
        F_y
    \end{bmatrix},
\end{equation}

where $m$ is the mass of the point. The collision avoidance constraint is defined similarly to \cite{Brito2019}:

\begin{equation}
    c_i^{j}=(\Delta \xi_{i,j})^T R(\theta_{i,j})^T H(\alpha,\beta) R(\theta_{i,j}) (\Delta \xi_{i,j})> 1,
\end{equation}

where $\Delta \xi_{i,j}$ is the distance between the center of $j$th ellipsoid and the vehicle at time $i$, $\alpha$, and $\beta$ are the semi-axes of an enlarged ellipse defined in \cite{Brito2019}. We use the model predictive controller defined in \cite{Lopez2018} where soft constraints are used for collision avoidance. 

\begin{mini}[2] 
{\xi, U}{J^t+J^u+J^c}
{}{}\label{mpc}
\addConstraint{(\xi_{i+1}-\xi_i)/dt=A\xi_i+Bu_i}{}
\addConstraint{c_i^j+\psi s_i>1}{}
\addConstraint{\xi_{min}\leq \xi_i \leq \xi_{max}}{}
\addConstraint{u_{min} \leq u_i \leq u_{max}}{}
\addConstraint{\xi_0=\xi_{init}}{}
\addConstraint{i=\{1,2,...,N\}}{},
\end{mini}

where $J^t$ is tracing cost, $J^u$ control cost, $s_k$ is slack variable, $J^c$ is the cost of the slack variable, $\psi$ is the sensitivity of relaxed collision constraint, $N$ is the prediction horizon, $\xi_{max},\xi_{min}$ are the state bounds, $u_{max},u_{min}$ are the control bounds and $\xi_{init}$ is the initial state. Interior Point Optimizer(IPOPT)\cite{ipopt} was used to solve the Nonlinear Programming(NLP) problem defined in \eqref{mpc}.

\section{RESULTS}

\begin{figure*}
    \centering
    \includegraphics[width=0.9\textwidth]{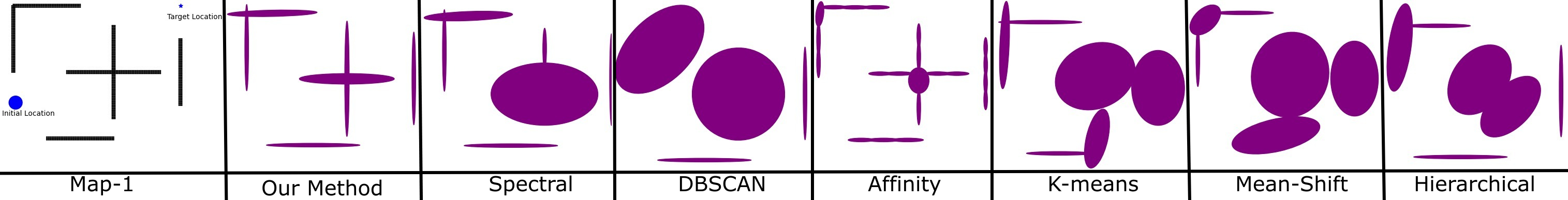} 
    \caption{Map-1. This scenario was created using 520 points where all obstacles are static.  }
    \label{fig:resstatic}
\end{figure*}

We test the proposed algorithm across five different scenarios. The parameters used in these scenarios are given in Table \ref{tab:parameters}. In the first scenario shown in \autoref{fig:resstatic} where there are only static obstacles, our algorithm generated six ellipsoids that closely represent the ground truth map. Affinity propagation\cite{Afinity} method also provided a suitable result, yet it generated 21 clusters and its computation time was more than five seconds, whereas our algorithm generates six ellipsoids in less than 100 milliseconds for the same map.

\begin{table}
    \centering
    \begin{tabular}{ |p{2cm}|p{3cm}|  }
     \hline
    Parameter & Value\\
     \hline
    $n_{ell_{max}}$ & 30 \\ 
    $e_{kyachiyan}$ & 0.05 \\
    $r_{ell_{threshold}}$ & 0.6 \\
    $\psi$ & 0.15 \\
    $N$ & 20 \\
    $Q$ & eye(4) \\
    $P$ & diag(.1,.1) \\
    $S$ & eye($n_{ell}$) \\
    $u_{max},u_{min}$ & 20,-20 N \\
    $\xi_{max}$ & $\infty,\infty,20,20$ m,m/s\\
    $\xi_{min}$ & $-\xi_{max}$ \\
    $m$ & 1 kg \\
     \hline
    \end{tabular}
    \caption{Parameters for the test. $S$ soft constraint penalty matrice, $P$ is the control penalty matrice and $Q$ is the tracking penalty matrice. }
    \label{tab:parameters}
\end{table}

\begin{figure}
    \centering
    \includegraphics[width=.3\textwidth]{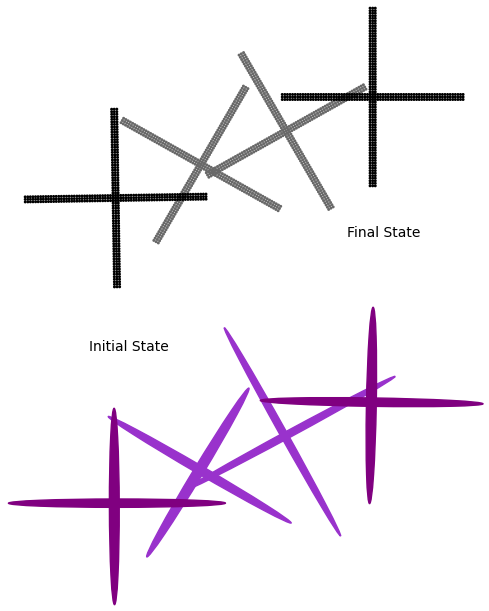}
    \caption{Map-2. This scenario which includes 214 points is generated to test motion estimation performance where the obstacle moves with the velocity of 5 m/s in the x direction, 2 m/s in the y direction, and rotates $\pi$/2 rad/s in the counter-clockwise direction. The result can be seen in \autoref{fig:resest}.}
    \label{fig:resmotion}
\end{figure}

\begin{figure}
    \centering
    \includegraphics[width=.45\textwidth]{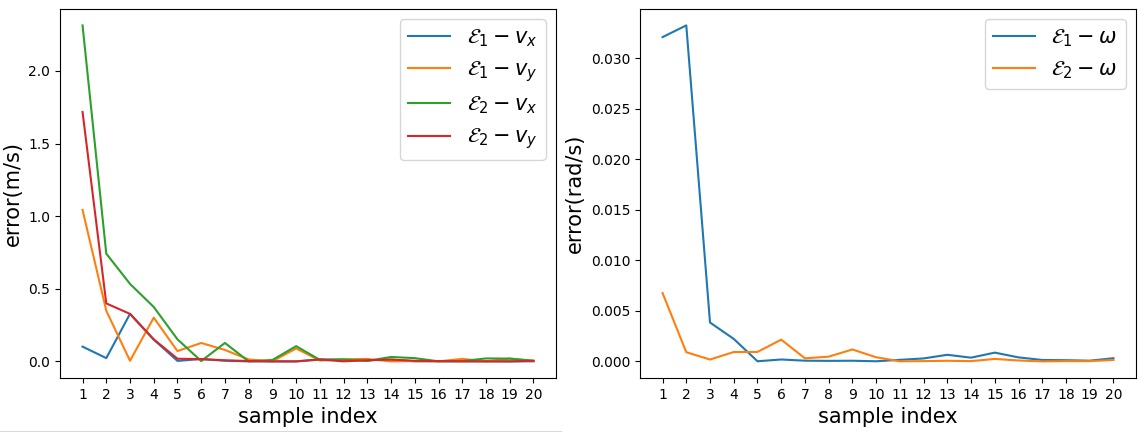}
    \caption{This figures show motion estimation errors in Map-2 seen in \autoref{fig:resmotion}. The left graph shows the error in linear velocity estimation and the right one is the error in angular velocity estimation.  The error is the absolute value of the difference between the real value and the estimated value.}
    \label{fig:resest}
\end{figure}

The second scenario presented in \autoref{fig:resmotion}, is designed to test motion estimation performance. A plus-shaped obstacle, with linear and angular velocities, was considered, and its motion was estimated via ellipsoids. The error in motion estimation is seen in \autoref{fig:resest} where the error becomes close to zero after a few samples.  

\begin{figure*}
    \centering
    \includegraphics[width=0.7\textwidth]{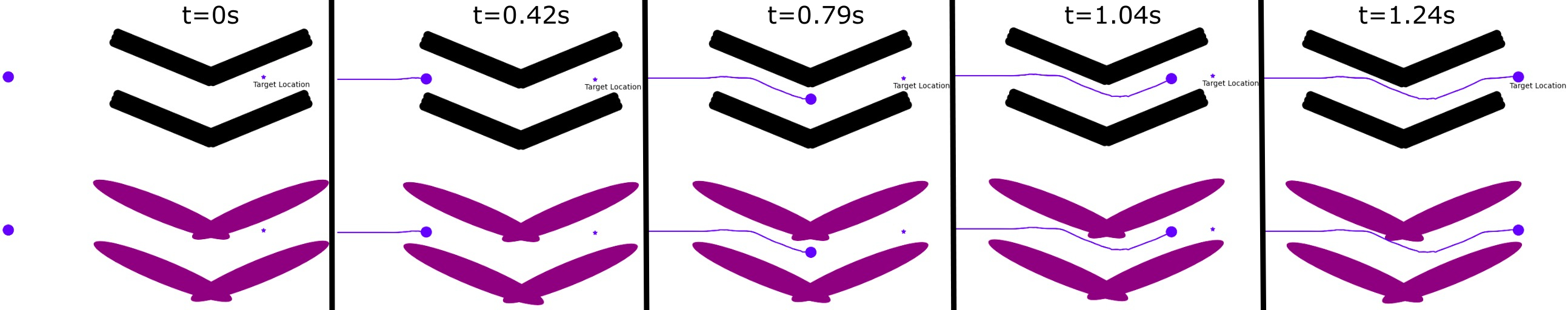} 
    \caption{Map-3. In this scenario, obstacles(248 points) are moving with a velocity of -5 m/s in the x(horizontal) direction and the vehicle is on the left side of the obstacle while the target location is on the right side of the obstacle.}
    \label{fig:linearmotion}
\end{figure*}

\begin{figure*}
    \centering
    \includegraphics[width=0.9\textwidth]{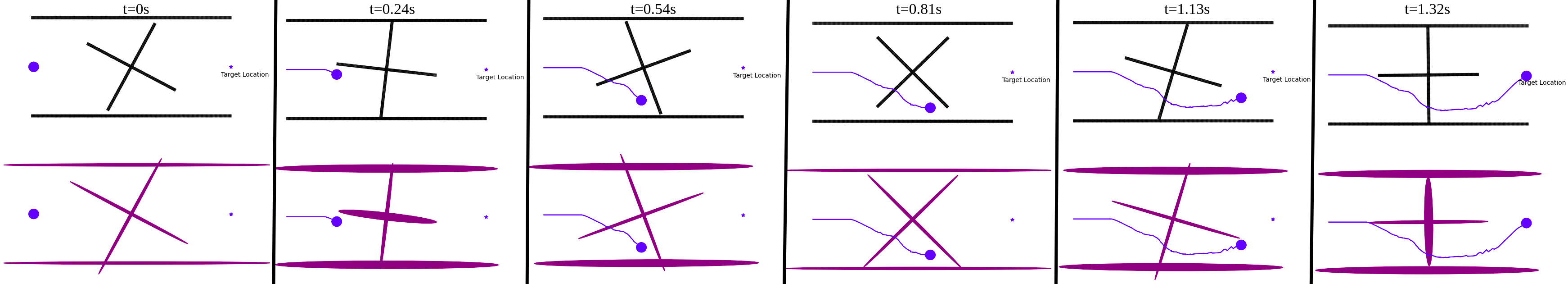} 
    \caption{Map-4. This scenario includes both static and dynamic obstacles where the total map is generated by using 908 points. While the walls on the upper and lower side do not move, the obstacle between the walls rotates with $\pi$/2 rad/s in the counter-clockwise direction.  }
    \label{fig:angularmotion}
\end{figure*}

\begin{figure*}
    \centering
    \includegraphics[width=0.9\textwidth]{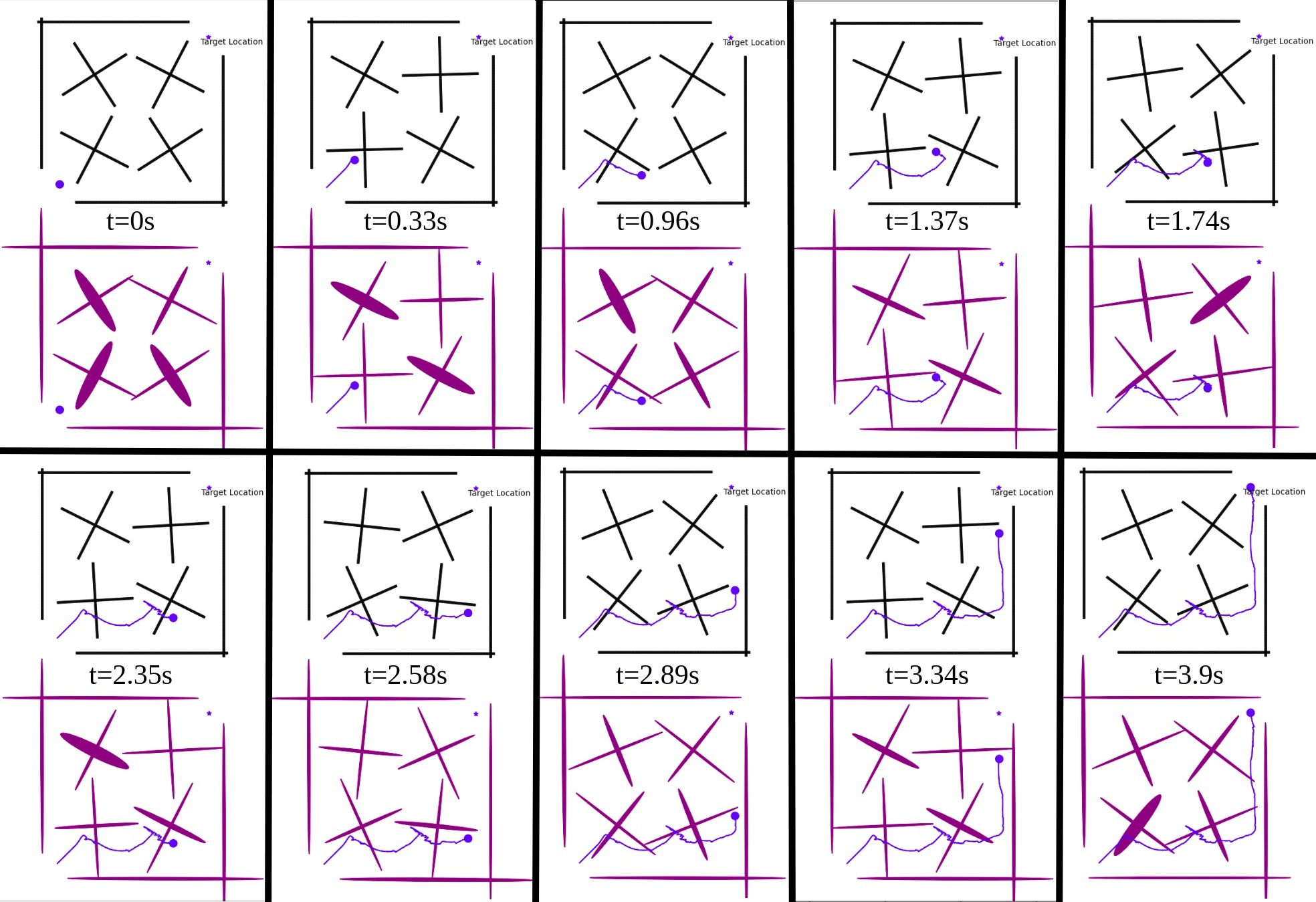} 
    \caption{Map-5. This scenario is generated by using 2418 points which include both static and dynamic obstacles.  The four walls are static and the four obstacles that are inside the walls rotate with $\pi$/2 rad/s  in the counter-clockwise direction.}
    \label{fig:complexmotion}
\end{figure*}

In the third scenario presented in \autoref{fig:linearmotion}, both obstacles are dynamic and move with constant velocity. The fourth scenario provided in \autoref{fig:angularmotion} involves two static and one dynamic obstacle, while the final scenario illustrated in \autoref{fig:complexmotion} involves four static and four dynamic obstacles. We combined some known clustering methods with the Khachiyan algorithm and test them in these scenarios. The GMM was not involved in this comparison due to the fact that it gives the result of our algorithm with the right number of clusters. For the algorithms that require the number of clusters, the optimal values which are: $6$ for Map-1,  $4$ for Map-3, $4$ for Map-4, and $12$ for Map-5 were used. Other parameters were tested and the most suitable ones were selected. The results are given in Table \ref{tab:sc3res}.

\begin{figure}
    \centering
    \includegraphics[width=.45\textwidth]{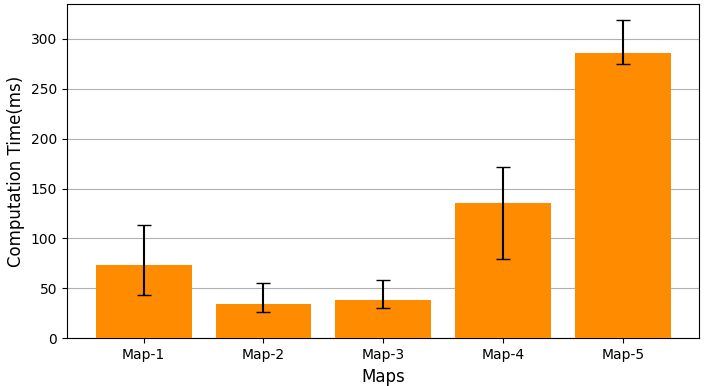}
    \caption{Computation time plot for each map. Map-1 includes 520 points, Map-2 includes 214 points, Map-3 includes 248 points, Map-4 includes 908 points and Map-5 includes 2418 points.}
    \label{fig:comptime}
\end{figure}

In the first scenario, The Spectral\cite{Spectral} method provided the best performance; however, the results were close to each other. Although the Spectral method provides convenient results when the number of clusters is chosen correctly, it beings to fail when the clusters overlap with each other. In scenario 3, our algorithm and the Spectral algorithm obtained the same result, whereas generated ellipsoids were also almost identical. In scenarios four and five, the other methods failed to reach the target location as a consequence of the wrong clustering. The blue plots in the figures show the motion generated by using our algorithm. All computations were run on a desktop computer with Intel core 7th generation i7 processor. The computation time of our algorithm in test maps is given in \autoref{fig:comptime}.

\begin{table}
    \centering
    \begin{tabular}{ |p{2cm}|p{1cm}|p{1cm}|p{1cm}|p{1cm}|  }
     \hline
     \multirow{2}{4em}{Method} & \multicolumn{4}{|c|}{Time to reach final location(s)} \\
     
     & Map-1 & Map-3 & Map-4 & Map-5\\
     \hline
    Ours & 0.96 & \textbf{1.24} & \textbf{1.32} & \textbf{3.9} \\
    Spectral$^*$ & \textbf{0.9} & \textbf{1.24} & - & -\\
    DBSCAN & 1.08 & 1.79 & - & -\\
    Affinity & 0.92 & 1.48 & - & -\\
    K-means$^*$ & 0.98 & - & - &-\\
    Mean-Shift & 0.97 & - & - & -\\
    Hierarchical$^*$ & 0.97 & - & - & -\\
     \hline
    \end{tabular}
    \caption{Planning comparison in custom maps. $^*$ means the algorithm requires the number of clusters and - means the vehicle did not reach the target location.}
    \label{tab:sc3res}
\end{table}

\section{CONCLUSION}
In this work, we propose an ellipsoid-based obstacle identification method to represent obstacles and estimate their motions without requiring the number of clusters beforehand. We validate our method in different scenarios that involve both static and dynamic obstacles including the rotating obstacles and compare our approach with other methods. In addition, we showed the motion estimation performances and the computational requirements of our algorithm. Our algorithm was able to produce accurate outputs even in environments with complex obstacles, outperforming the compared approaches.

\newpage

\bibliographystyle{IEEEtran}
\bibliography{references}

\end{document}